\pdfoutput=1
\documentclass{article}

\usepackage[preprint,nonatbib]{nips_2018}

\usepackage[utf8]{inputenc} 
\usepackage[T1]{fontenc}    
\usepackage{hyperref}       
\usepackage{url}            
\usepackage{booktabs}       
\usepackage{amsfonts}       
\usepackage{nicefrac}       
\usepackage{microtype}      
\usepackage[pdftex]{graphicx} 
\usepackage{rotating}
\usepackage{colortbl}
\usepackage{color}
\title{ML-Schema: Exposing the Semantics of Machine Learning with Schemas and Ontologies}

\author{Gustavo Correa Publio \\ AKSW Group\\ University of Leipzig, Germany
	\And
	Diego Esteves
	\\ SDA Research\\ University of Bonn, Germany 
	\And
	Agnieszka \L{}awrynowicz \\
 Poznan University of Technology\\Poland 
 \And
  Pan\v{c}e Panov \\
   Jo\v{z}ef Stefan Institute\\Ljubljana, Slovenia
  \And
  Larisa Soldatova \\
  Brunel University, UK \\
 \And
 Tommaso Soru\\ AKSW Group\\ University of Leipzig, Germany
  \And
  Joaquin Vanschoren \\
  Eindhoven \\University of Technology\\The Netherlands 
  \And
	Hamid Zafar
	\\ SDA Research\\ University of Bonn, Germany
}

\begin{document}

\maketitle

\begin{abstract}
The ML-Schema, proposed by the W3C Machine Learning Schema Community Group, is a top-level ontology that provides a set of classes, properties, and restrictions for representing and interchanging information on machine learning algorithms, datasets, and experiments. It can be easily extended and specialized and it is also mapped to other more domain-specific ontologies developed in the area of machine learning and data mining. In this paper we overview existing state-of-the-art machine learning interchange formats and present the first release of ML-Schema, a canonical format resulted of more than seven years of experience among different research institutions. We argue that exposing semantics of machine learning algorithms, models, and experiments through a canonical format may pave the way to better interpretability and to realistically achieve the full interoperability of experiments regardless of platform or adopted workflow solution.


\end{abstract}

\section{Introduction}
Complex machine learning models have recently achieved great successes in many predictive tasks. Despite their successes, a major problem is that they are often hard to interpret, which may affect their safeness and the level of trust of their users. The problem of interpretability is one of the key research issues in the area of knowledge engineering and the Semantic Web community, which deals with making the semantics of various phenomena explicit. In this community, the problem of dealing with trust and traceability has gained a major interest last years and resulted in proposals and uptake of models such as the provenance ontology PROV-O~\cite{lebo2013prov}.

\begin{figure}[t]
	\centering
	\includegraphics[width=0.7\linewidth]{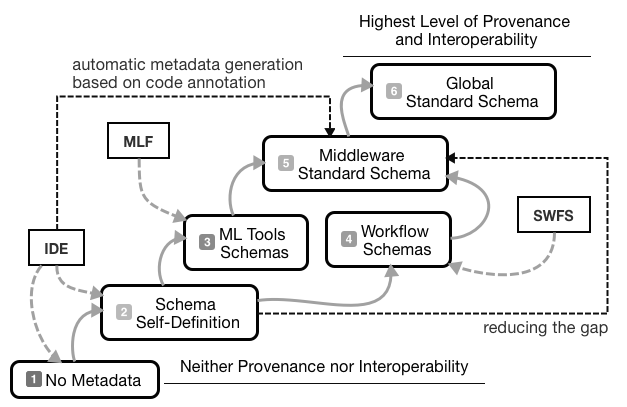}
	\caption{Vertical and Horizontal Interoperability across ML Environments.}
	\label{fig:metalevel} 
\end{figure}

Despite recent efforts to achieve a high level of interoperability of ML experiments through existing workflow systems~\cite{Tcheremenskaia:2012}, metadata repositories~\cite{vanschoren2014openml,neto2016wasota} and schemata~\cite{DBLP:conf/i-semantics/EstevesMNSUAL15,DBLP:journals/datamine/PanovSD14,DBLP:journals/ws/KeetLdKNPSH15}, we still run into problems created due to the existence of different ML platforms: each of
those has a specific conceptualization or schema for representing data and metadata. Figure \ref{fig:metalevel} depicts the complexity of this scenario: (1) and (2) representing the worse scenario to achieve reproducibility of the experiments; (3) and (4) which - although defining a known (local) schema - lack interoperability, since they follow different standards. To reduce this gap, the aforementioned ML vocabularies and ontologies have been proposed (e.g., MEX and Expose) (5); and finally (6) which encompasses the higher level of interoperability to allow fully reproducible experiments.  


In line with those efforts to bridge the gap, in this paper, we present the ML-Schema\footnote{ML-Schema: \url{http://ml-schema.github.io/documentation/}}~\cite{mlschema:201610}, developed within the W3C Machine Learning Schema Community Group\footnote{W3C ML-Schema Community Group: \url{https://www.w3.org/community/ml-schema/}}. It is a simply shared schema that provides a set of classes, properties, and restrictions that can be used to represent and interchange information on Machine Learning (ML) algorithms, datasets, and experiments. It can be easily specialized to create new classes and properties and it is also mapped to more specific ontologies and vocabularies on machine learning \cite{DBLP:conf/i-semantics/EstevesMNSUAL15,DBLP:journals/ml/VanschorenBPH12,DBLP:journals/datamine/PanovSD14,DBLP:journals/ws/KeetLdKNPSH15}, for instance to represent Deep Learning problems, which are naturally harder to design when compared to supervised approaches. These ontologies, in turn, contain terms for representing more detailed characteristics and properties of ML datasets, algorithms, models, and experiments. 

Ultimately, we believe that involving such a canonical and standardized model meta-data descriptors in design of interpretable methods for ML may lead to better insights into the data and the properties of ML algorithms.

The gap can be further significantly reduced by achieving interoperability among state-of-the-art (SOTA) schemata of those resources (Figure \ref{fig:metalevel}: item 5), i.e., achieving the horizontal interoperability (Figure \ref{fig:metalevel}: item 6).
Therefore, different groups of researchers could exchange SOTA metadata files in a transparent manner via web services, e.g.: from OntoDM and MEX (\texttt{MLSchema.Schema data = mlschema.convert(``myFile.ttl'', MLSchema.Ontology.OntoDM, MLSchema.Ontology.MEX)}).
Furthermore, the canonical format also directly benefits different environments, such as ML ecosystems (e.g. OpenML \cite{vanschoren2014openml}) and ML Metadata Repositories (e.g. WASOTA \cite{neto2016wasota}) which can benefit on the mappings of a shared standard.

\section{The ML-Schema}
\label{sec:mlschema}
\begin{figure}[t]
  \centering
  \includegraphics[width=0.9\linewidth]{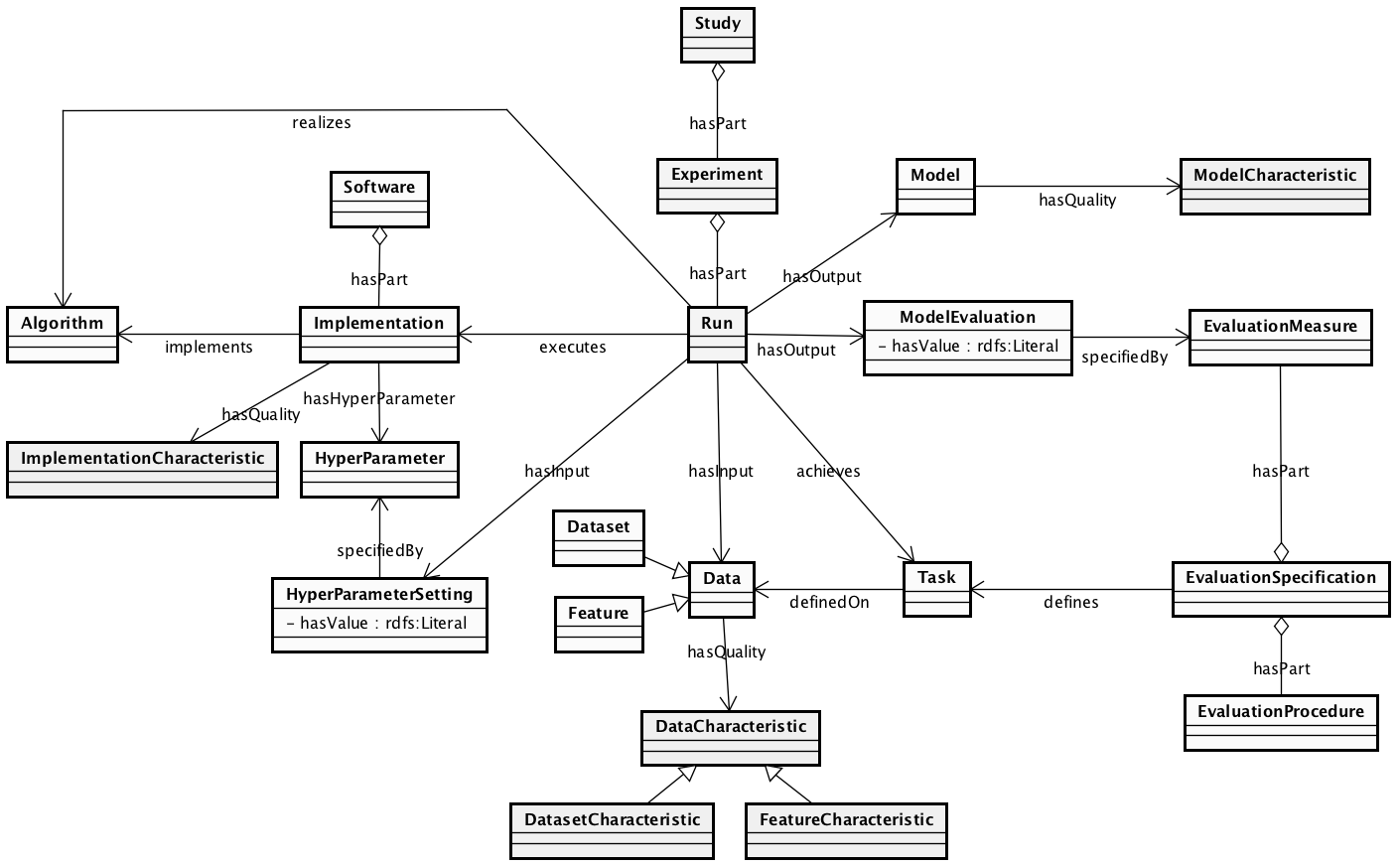}
  \caption{ML-Schema Core classes. Boxes represent classes. Arrows without filled heads represent properties, arrows with empty heads represent subclass relations, and arrows with diamonds represent part-of relations. \label{fig:mlschema}} 
\end{figure}

In Fig.~\ref{fig:mlschema}, we depict the classes and the relationships between the classes representing the ML-Schema. The schema contains classes for representing different aspects of machine learning. This includes representations data, datasets and data/dataset characteristics. Next, it includes representations of algorithms, implementations, parameters of implementations and software. Furthermore, the schema contains representations of models, model characteristics and model evaluations. Finally, the schema has the ability to represent machine learning experiments with different granularity. This includes representations of runs of implementations of algorithms on the lowest level and representation of studies at the highest level. 

\section{Machine Learning ontologies}
\label{sec:ontologies}

In this section, we present the relationship of the ML-Schema to other proposed ontologies, and vocabularies for the domain of machine learning. 
The development of ML-Schema was highly influenced from, initially independent, research  of  several  groups  on  modeling the domain of machine learning.  
Due to this the classes and relations present in the ML-Schema  re-appear in the current ML ontologies and vocabularies. 
In Table~\ref{tab:ontos}, we present the mapping between the terms present in the ML-Schema and the current ML ontologies and vocabularies. Below, we describe each of the mentioned ontology/vocabulary.

\paragraph{The OntoDM-core ontology.}

The OntoDM-core ontology has been designed to provide generic representations of principle entities in the area of data mining~\cite{DBLP:journals/datamine/PanovSD14}. In one of the preliminary versions of the ontology, the authors decided to align the proposed ontology with the Ontology of Biomedical Investigations (OBI) \cite{10.1371/journal.pone.0154556} and consequently with the Basic Formal Ontology (BFO) \cite{Arp:2015:BOB:2846229} at the top level, in terms of top-level classes and the set of relations. That was beneficial for structuring the domain in a more elegant way and the basic differentiation of information entities, implementation entities and processual entities. In this context, the authors proposed a horizontal description structure that includes three layers: a specification layer, an implementation layer, and an application layer. The specification layer in general contains information entities. In the domain of data mining, example classes are data mining task and data mining algorithm. The implementation layer in general contains qualities and entities that are realized in a process, such as parameters and implementations of algorithms. The application layer contains processual classes, such as the execution of the data mining algorithm. 	

\paragraph{The Expos\'e ontology.}

The main goal of Expos\'e~\cite{DBLP:journals/ml/VanschorenBPH12} is to describe (and reason about) machine learning experiments. It is built on top of OntoDM, as well as top-level ontologies from bio-informatics. It is currently used in OpenML\footnote{OpenML: \url{http://www.openml.org/}}, as a way to structure data (e.g. database design) and share data (APIs). MLSchema will be used to export all OpenML data as linked open data (in RDF).

\paragraph{The DMOP ontology.}

The Data Mining OPtimization Ontology (DMOP)~\cite{DBLP:journals/ws/KeetLdKNPSH15} has been developed with a primary use case in meta-mining, that is meta-learning extended to an analysis of full DM processes. At the level of both single algorithms and more complex workflows, it follows a very similar modeling pattern as described in the MLSchema. To support meta-mining, DMOP contains a taxonomy of algorithms used in DM processes which are described in detail in terms of their underlying assumptions, cost functions, optimization strategies, generated models or pattern sets, and other properties. Such a ``glass box'' approach which makes explicit internal algorithm characteristics allows meta-learners using DMOP to generalize over algorithms and their properties, including those algorithms which were not used for training meta-learners.
DMOP also contains sub-taxonomies of ML models and provides vocabulary to describe their properties and characteristics, e.g. model structures, model complexity measures, parameters.

\paragraph{The MEX vocabulary.}

The MEX vocabulary~\cite{DBLP:conf/i-semantics/EstevesMNSUAL15} has been designed to reuse existing ontologies (e.g., PROV-O~\cite{lebo2013prov})  for representing basic machine learning experiment configuration and its outcomes. The aim is not to describe a complete data-mining process, which can be modeled by more complex and semantically refined structures. Instead, MEX is designed
to provide a simple and lightweight vocabulary for exchanging basic machine learning metadata in order to achieve
a high level of interoperability. Moreover, the schema aims to serve as a basis for data management of ML outcomes in the context of WASOTA \cite{neto2016wasota}. The principal components are: \texttt{mex-algo}\footnote{\url{http://mex.aksw.org/mex-algo}} which describes \textit{algorithms} and \textit{hyperparameters}, \texttt{mex-core}\footnote{\url{http://mex.aksw.org/mex-core}} which is the basis to describe an \textit{experiment}, its \textit{configurations} and \textit{executions} and \texttt{mex-perf}\footnote{\url{http://mex.aksw.org/mex-perf}}, the layer to map the \textit{outcomes} (i.e., performance measures).

\begin{table}
 \tiny
  \caption{Mapping between the terms between the ML-Schema and the different ML/DM ontologies and vocabularies}
  \label{tab:ontos}

  \centering
  \begin{tabular}{l l lll}
   
\toprule
\textbf{ML-Schema} & OntoDM-core & DMOP  & Expose & MEX Vocabulary 
\\
\midrule
\cellcolor[gray]{0.8}Task &
Data mining task &
DM-Task &
Task &
mexcore:ExperimentConfiguration
\\
\cellcolor[gray]{0.8}Algorithm&
Data mining algorithm&
DM-Algorithm&
Algorithm&
mexalgo:Algorithm
\\
\cellcolor[gray]{0.8}Software&
Data mining software&
DM-Software&
N/A&
mexalgo:Tool
\\
\cellcolor[gray]{0.8}Implementation&
Data mining algorithm &
DM-Operator&
Algorithm implementation&
N/A
\\
\cellcolor[gray]{0.8}HyperParameter&
Parameter&
Parameter&
Parameter&
mexalgo:HyperParameter
\\
\cellcolor[gray]{0.8}HyperParameterSetting&
Parameter setting&
OpParameterSetting&
Parameter setting&
N/A
\\
\cellcolor[gray]{0.8}Study&
Investigation&
N/A&
N/A&
mexcore:Experiment
\\
\cellcolor[gray]{0.8}Experiment&
N/A&
DM-Experiment&
Experiment&
N/A
\\
\cellcolor[gray]{0.8}Run&
Data mining alg. execution&
DM-Operation&
Algorithm execution&
mexcore:Execution
\\
\cellcolor[gray]{0.8}Data&
Data item&
DM-Data&
N/A&
mexcore:Example
\\
\cellcolor[gray]{0.8}Dataset&
DM dataset&
DataSet&
Dataset&
mexcore:Dataset
\\
\cellcolor[gray]{0.8}Feature&
N/A&
Feature&
N/A&
mexcore:Feature
\\
\cellcolor[gray]{0.8}DataCharacteristic&
Data specification&
DataCharacteristic&
Dataset specification&
N/A
\\
\cellcolor[gray]{0.8}DatasetCharacteristic&
Dataset specification&
DataSetCharacteristic&
Data quality&
N/A
\\
\cellcolor[gray]{0.8}FeatureCharacteristic&
Feature specification&
FeatureCharacteristic&
N/A&
N/A
\\
\cellcolor[gray]{0.8}Model&
Generalization&
DM-Hypothesis &
Model&
mexcore:Model
\\
\cellcolor[gray]{0.8}ModelCharacteristic&
Generalization quality&
HypothesisCharacteristic&
Model Structure, Parameter, ...&
N/A
\\
\cellcolor[gray]{0.8}ModelEvaluation&
Generalization evaluation&
ModelPerformance&
Evaluation&
N/A
\\
\cellcolor[gray]{0.8}EvaluationMeasure&
Evaluation datum&
ModelEvaluationMeasure&
Evaluation measure&
mexperf:PerformanceMeasure
\\
\cellcolor[gray]{0.8}EvaluationSpecification&
N/A&
N/A&
N/A&
N/A
\\
\cellcolor[gray]{0.8}EvaluationProcedure&
Evaluation algorithm&
ModelEvaluationAlgorithm&
Performance Estimation&
N/A\\
 \bottomrule
  \end{tabular}
\end{table}

\section{Conclusions}
In this extended abstract, we have presented ML-Schema, a lightweight schema for modeling ML domain. 
The ML-Schema aligns more fine-grained ontologies and vocabularies, some of which contain detailed vocabulary for representing meta-data on ML models.
The vocabulary and axiomatization included in those resources may be used to make explicit the semantics of ML models, making them better interpretable for human users. 

\subsubsection*{Acknowledgments}
Gustavo Correa Publio acknowledges the support of the Smart
Data Web BMWi project (GA-01MD15010B) and CNPq Foundation (201808/2015-3). Agnieszka \L{}awrynowicz acknowledges the support from the National Science Centre,
Poland, within the grant number 2014/13/D/ST6/02076. Pan\v{c}e Panov acknowledges the support of the Slovenian Research Agency within the grant J2-9230. Hamid Zafar acknowledges the EU H2020 grants for the WDAqua (GA no. 642795) project.


\newpage
\small
\bibliographystyle{plain}
\bibliography{main}

\begin{thebibliography}{10}

\bibitem{Arp:2015:BOB:2846229}
Robert Arp, Barry Smith, and Andrew~D. Spear.
\newblock {\em Building Ontologies with Basic Formal Ontology}.
\newblock The MIT Press, 2015.

\bibitem{10.1371/journal.pone.0154556}
Anita Bandrowski, Ryan Brinkman, Mathias Brochhausen, Matthew~H. Brush, Bill
  Bug, Marcus~C. Chibucos, Kevin Clancy, Mélanie Courtot, Dirk Derom, Michel
  Dumontier, Liju Fan, Jennifer Fostel, Gilberto Fragoso, Frank Gibson,
  Alejandra Gonzalez-Beltran, Melissa~A. Haendel, Yongqun He, Mervi Heiskanen,
  Tina Hernandez-Boussard, Mark Jensen, Yu~Lin, Allyson~L. Lister, Phillip
  Lord, James Malone, Elisabetta Manduchi, Monnie McGee, Norman Morrison,
  James~A. Overton, Helen Parkinson, Bjoern Peters, Philippe Rocca-Serra, Alan
  Ruttenberg, Susanna-Assunta Sansone, Richard~H. Scheuermann, Daniel Schober,
  Barry Smith, Larisa~N. Soldatova, Christian~J. Stoeckert, Jr., Chris~F.
  Taylor, Carlo Torniai, Jessica~A. Turner, Randi Vita, Patricia~L. Whetzel,
  and Jie Zheng.
\newblock The ontology for biomedical investigations.
\newblock {\em PLoS ONE}, 11(4):1--19, 04 2016.

\bibitem{mlschema:201610}
Diego Esteves, Agnieszka Lawrynowicz, Pance Panov, Larisa~N. Soldatova, Tommaso
  Soru, and Joaquin Vanschoren.
\newblock Ml schema core specification.
\newblock Technical report, W3C, 10 2016.
\newblock http://www.w3.org/2016/10/mls/.

\bibitem{DBLP:conf/i-semantics/EstevesMNSUAL15}
Diego Esteves, Diego Moussallem, Ciro~Baron Neto, Tommaso Soru, Ricardo Usbeck,
  Markus Ackermann, and Jens Lehmann.
\newblock {MEX} vocabulary: a lightweight interchange format for machine
  learning experiments.
\newblock In {\em Proceedings of the 11th International Conference on Semantic
  Systems, {SEMANTICS} 2015, Vienna, Austria, September 15-17, 2015}, pages
  169--176, 2015.

\bibitem{DBLP:journals/ws/KeetLdKNPSH15}
C.~Maria Keet, Agnieszka Lawrynowicz, Claudia d'Amato, Alexandros Kalousis,
  Phong Nguyen, Ra{\'{u}}l Palma, Robert Stevens, and Melanie Hilario.
\newblock The data mining optimization ontology.
\newblock {\em J. Web Sem.}, 32:43--53, 2015.

\bibitem{lebo2013prov}
Timothy Lebo, Satya Sahoo, Deborah McGuinness, Khalid Belhajjame, James Cheney,
  David Corsar, Daniel Garijo, Stian Soiland-Reyes, Stephan Zednik, and Jun
  Zhao.
\newblock Prov-o: The prov ontology.
\newblock {\em W3C Recommendation}, 30, 2013.

\bibitem{neto2016wasota}
Ciro~Baron Neto, Diego Esteves, Tommaso Soru, Diego Moussallem, Andre
  Valdestilhas, and Edgard Marx.
\newblock Wasota: What are the states of the art?
\newblock In {\em SEMANTiCS (Posters, Demos, SuCCESS)}, 2016.

\bibitem{DBLP:journals/datamine/PanovSD14}
Pance Panov, Larisa~N. Soldatova, and Saso Dzeroski.
\newblock Ontology of core data mining entities.
\newblock {\em Data Min. Knowl. Discov.}, 28(5-6):1222--1265, 2014.

\bibitem{Tcheremenskaia:2012}
Olga Tcheremenskaia, Romualdo Benigni, Ivelina Nikolova, Nina Jeliazkova,
  Sylvia~E Escher, Monika Batke, Thomas Baier, Vladimir Poroikov, Alexey
  Lagunin, Micha Rautenberg, and Barry Hardy.
\newblock Opentox predictive toxicology framework: toxicological ontology and
  semantic media wiki-based opentoxipedia.
\newblock {\em Journal of Biomedical Semantics}, 2012.

\bibitem{DBLP:journals/ml/VanschorenBPH12}
Joaquin Vanschoren, Hendrik Blockeel, Bernhard Pfahringer, and Geoffrey Holmes.
\newblock Experiment databases - {A} new way to share, organize and learn from
  experiments.
\newblock {\em Machine Learning}, 87(2):127--158, 2012.

\bibitem{vanschoren2014openml}
Joaquin Vanschoren, Jan~N Van~Rijn, Bernd Bischl, and Luis Torgo.
\newblock Openml: networked science in machine learning.
\newblock {\em ACM SIGKDD Explorations Newsletter}, 15(2):49--60, 2014.

\end{thebibliography}

\end{document}